\documentclass[sigconf]{acmart}
\AtBeginDocument{%
  }

\copyrightyear{2026}
\acmYear{2026}
\setcopyright{cc}
\setcctype{by}
\acmConference[MM '26]{Proceedings of the 34th ACM International Conference on Multimedia}{November 10--14, 2026}{Rio de Janeiro, Brazil}
\acmBooktitle{Proceedings of the 34th ACM International Conference on Multimedia (MM '26), November 10--14, 2026, Rio de Janeiro, Brazil}
\acmDOI{10.1145/3767308.3835088}
\acmISBN{979-8-4007-2213-4/2026/11}
\acmSubmissionID{1071}

\usepackage{booktabs}     
\usepackage{multirow}     
\usepackage{colortbl}     
\usepackage{xcolor}       
\usepackage{graphicx}
\usepackage{makecell}
\usepackage{pifont}
\newcommand{\cmark}{\ding{51}} 
\newcommand{\xmark}{\ding{55}} 
\usepackage{algorithm}
\usepackage{algorithmic}
\usepackage{xcolor} 
\usepackage{balance} 

\begin{document}

\title[Token-Budget Distillation]{Token-Budget Distillation: Transferring Full-Token Semantics to Compressed Video Vision-Language Models}
\author{Xiaoyang Guo}
\orcid{0009-0001-0848-1369}
\affiliation{
\institution{Sun Yat-sen University}
\city{Guangzhou}
\state{Guangdong}
\country{China}
}
\email{guoxy77@mail2.sysu.edu.cn}

\author{Guoping Luo}
\orcid{0009-0003-8283-6273}
\affiliation{
\institution{The University of British Columbia}
\city{Vancouver}
\state{British Columbia}
\country{Canada}
}
\email{luoguoping2022@163.com}

\author{Jusheng Zhang}
\orcid{0009-0004-5743-2936}
\affiliation{
\institution{Sun Yat-sen University}
\city{Guangzhou}
\state{Guangdong}
\country{China}
}
\email{jushengzhang88889@gmail.com}

\author{Keze Wang}
\correspondingauthor
\authornote{Corresponding authors.}
\orcid{0000-0002-7817-8306}
\affiliation{
\institution{Sun Yat-sen University}
\city{Guangzhou}
\state{Guangdong}
\country{China}
}
\email{kezewang@gmail.com}

\author{Wenhao Wang}
\correspondingauthor
\authornotemark[1]
\orcid{0000-0001-8727-1572}
\affiliation{
\institution{Vast Intelligence Lab}
\city{Sydney}
\state{New South Wales}
\country{Australia}
}
\email{wangwenhao@vastilab.com}
\begin{abstract}
    Adapting video vision-language models (VLMs) is computationally expensive because video inputs produce a large number of visual tokens, making both fine-tuning and inference costly. Although visual token compression can reduce this overhead, direct adaptation on compressed inputs often causes semantic drift and noticeable performance degradation. We present Token-Budget Distillation (TBD), a parameter-efficient fine-tuning framework for adapting video VLMs under a fixed token budget. TBD freezes the pretrained backbone, updates only LoRA adapters, and integrates FlashVID-based visual token compression into the video pathway. To preserve full-token semantics under compression, TBD employs a dual-path teacher-student design, where a full-token teacher provides stable supervision and a compressed student is optimized with task loss, answer-region KL distillation, GT-anchored margin distillation, and reliability-aware KD control. This design enables the student to recover the semantic behavior of the full-token model while remaining efficient under aggressive token reduction. We evaluate TBD on three video VLM backbones, including LLaVA-Video, LLaVA-OneVision, and Qwen3-VL-8B-Instruct, across four video understanding benchmarks. TBD consistently outperforms compression-only baselines under both moderate and aggressive compression. On LLaVA-Video at retention ratio R=10\%, TBD preserves 97.0\% of the Vanilla model's average accuracy; on LLaVA-OneVision at R=10\%, it achieves an average score of 58.4 and matches 100.0\% relative accuracy. 
\end{abstract}




\maketitle

\section{Introduction}

Recent advances in video vision-language models (VLMs) have led to strong performance on video question answering, instruction following, and long-video understanding tasks \cite{maaz2024video,lin2024video,zhang2024llava,li2024llava,wang2024qwen2,bai2025qwen3,3DAgent,MAB,KABB}. Representative systems such as Video-ChatGPT \cite{maaz2024video}, Video-LLaVA \cite{lin2024video}, LLaVA-OneVision \cite{li2024llava}, LLaVA-Video \cite{zhang2024llava}, and the Qwen-VL \cite{wang2024qwen2, bai2025qwen3} family demonstrate that large multimodal backbones can reason over spatiotemporal visual content when provided with rich video representations. However, adapting these models remains computationally prohibitive. Compared with static images, videos trigger a "token explosion" across frames, causing the cost of both fine-tuning and inference to scale with sequence length. 

While visual token compression has emerged as a promising way to mitigate this overhead, existing methods predominantly treat compression as a post-hoc inference accelerator. Recent token pruning and merging methods show that a large fraction of visual tokens can be removed or merged with limited loss in accuracy \cite{bolya2022token,chen2024image,yang2025visionzip,GAM,CF,MAT}. This idea has also been extended to video settings through methods such as PruneVid \cite{huang2025prunevid}, FrameFusion \cite{fu2024framefusion}, FastVID \cite{shen2025fastvid}, and FlashVID \cite{fan2026flashvid}, which reduce spatiotemporal redundancy to improve efficiency. In particular, training-free video compression methods are attractive because they can be plugged into existing video VLMs without retraining the compression module itself \cite{fan2026flashvid,MMCOT,HTC,OSC,zhang2026failuredriven,zhang2026kolmogorovarnoldfouriernetworks}. When fine-tuning is inherently required under a strict token budget, directly applying task supervision on compressed inputs exposes a critical vulnerability: \textbf{nontrivial semantic drift}. Specifically, the indiscriminate merging or dropping of spatiotemporal tokens often dilutes fine-grained foreground actions into aggregated background noise. Consequently, the student model's visual attention drifts from precise, discriminative actions (e.g., "Riding a bicycle") towards ambiguous, generalized conceptual representations (e.g., "Moving object"). This leads to degraded confidence and a significant loss of the full-token model's original semantic reasoning capability.

\begin{figure*}[h]
  \centering
  \includegraphics[width=\linewidth]{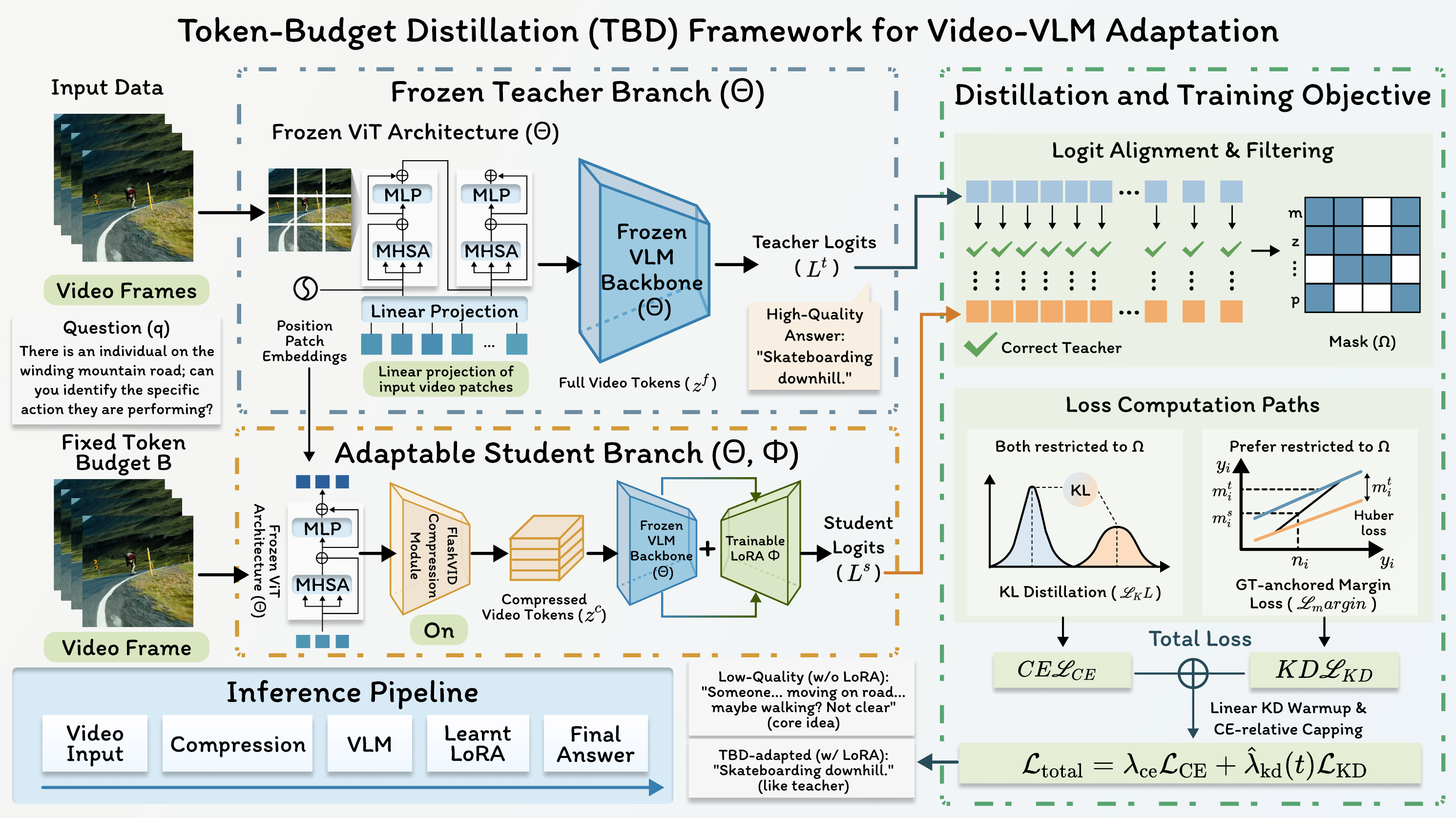}
  \caption{Overview of the Token-Budget Distillation (TBD) framework. A full-token teacher branch provides reliable guidance, while a compressed student branch with LoRA is optimized under a fixed token budget.}
  \Description{Block diagram of the TBD training framework. A video and a question are processed by two branches sharing one frozen backbone: a teacher branch that consumes the full visual token sequence, and a student branch in which FlashVID compresses the tokens before the language model while LoRA adapters are trained. The output logits of the two branches are compared to compute the task cross-entropy loss, the answer-region KL distillation loss, and the GT-anchored margin loss, and only the LoRA parameters receive gradient updates.}
  \label{fig:pipeline}
\end{figure*}

To bridge this gap, we propose \textbf{Token-Budget Distillation (TBD)}, a parameter-efficient fine-tuning framework that shifts the paradigm from simple compression-aware tuning to \textit{teacher-guided semantics preservation}. As detailed in the dual-branch architecture in Figure \ref{fig:pipeline}, TBD maps the robust knowledge of a full-token teacher into a compressed student operating under a stringent token budget. TBD freezes the pretrained backbone and updates only LoRA adapters \cite{hu2022lora}, integrating FlashVID-based compression into the forward pathway. To ensure stable transfer despite severe token reduction, we formulate a reliable distillation module that synergizes answer-region KL distillation \cite{hinton2015distilling,DAAPO}, ground-truth-anchored margin distillation, and dynamic KD warmup.

TBD goes beyond macro-level accuracy to mechanistically rehabilitate the compressed model's internal representations. Rather than treating the VLM as a black box, we counteract semantic drift through targeted constraints computed within our distillation module. Answer-region KL alignment ensures the student only mimics the teacher's most confident and relevant token distributions, filtering out noise and pulling the student's conceptual embedding back to the teacher's manifold. Simultaneously, the GT-anchored margin distillation enforces a robust confidence gap between the target response and competing tokens.

We evaluate TBD on three video VLM backbones---LLaVA-Video \cite{zhang2024llava}, LLaVA-OneVision \cite{li2024llava}, and Qwen3-VL-8B-Instruct \cite{bai2025qwen3}---across four video understanding benchmarks: MVBench \cite{li2024mvbench}, VideoMME \cite{fu2025video}, EgoSchema \cite{mangalam2023egoschema}, and LongVideoBench \cite{wu2024longvideobench}. TBD consistently and substantially outperforms compression-only baselines under both moderate ($R=20\%$) and aggressive ($R=10\%$) compression regimes. Notably, on LLaVA-Video at an aggressive $R=10\%$ retention ratio, TBD preserves 97.0\% of the Vanilla model's average accuracy; on LLaVA-OneVision at $R=10\%$, it achieves an average score of 58.4, fully matching 100.0\% relative accuracy. 

Our contributions are summarized as follows:
\begin{itemize}
    \item \textbf{Identification of Semantic Drift:} We pinpoint and formalize the semantic drift pathology caused by direct fine-tuning on compressed visual tokens, demonstrating how spatial dilution leads to cognitive ambiguity, thus reframing efficient adaptation as a semantics preservation challenge.
    \item \textbf{Token-Budget Distillation Framework:} We propose TBD, a compression-aware LoRA adaptation framework. As illustrated in Figure \ref{fig:pipeline}, by integrating answer-region KL alignment, GT-anchored margin supervision, and reliability-aware KD control, TBD safely transfers full-token semantics to a severely budgeted student.
    \item \textbf{Deep Mechanistic Insights \& Empirical Success:} TBD achieves consistent state-of-the-art accuracy-efficiency trade-offs across three major video VLM backbones. Beyond macro-metrics, our analyses show that TBD rectifies output distribution misalignment, reconstructs robust decision boundaries, and arrests visual attention drift.
\end{itemize}

\section{Related Work}

\paragraph{Video vision-language models}
Recent video vision-language models extend large multimodal architectures from images to videos and have rapidly improved performance on video understanding tasks. Early representative efforts, including VideoChat \cite{li2025videochat}, Valley \cite{luo2023valley}, Video-LLaMA \cite{zhang2023video}, Video-ChatGPT \cite{maaz2024video}, and Video-LLaVA \cite{lin2024video}, demonstrated that large language models can be coupled with video encoders for conversational, instruction-following, and temporally aware video understanding. Subsequent works further explored long-video memory, compact visual representation, temporal grounding, and efficient transfer from image-based backbones, as in MovieChat \cite{song2024moviechat}, LLaMA-VID \cite{li2024llama}, VTimeLLM \cite{huang2024vtimellm}, PLLaVA \cite{xu2024pllava}, ST-LLM \cite{liu2024st}, Video-STaR \cite{zohar2024video}, and VideoLLaMA 2 \cite{cheng2024videollama}. More recent models, such as LLaVA-OneVision \cite{li2024llava}, LLaVA-Video \cite{zhang2024llava}, Qwen2-VL \cite{wang2024qwen2}, and Qwen3-VL \cite{bai2025qwen3}, further strengthen video reasoning through improved architectural design and instruction-tuning data. Despite their differences in spatiotemporal modeling, long-context handling, and positional encoding, these models share the same computational bottleneck: the number of visual tokens grows rapidly with video length. This inevitably leads to quadratic complexity in self-attention mechanisms, making both fine-tuning adaptation and real-time inference increasingly expensive. This critical scaling behavior motivates the problem setting studied in our work.

\paragraph{Visual token compression for video VLMs}
To reduce the computational burden of long visual sequences, a large body of work studies visual token reduction for multimodal models. Representative generic pruning and merging methods include ToMe \cite{bolya2022token}, FastV \cite{chen2024image}, PyramidDrop \cite{xing2025pyramiddrop}, SparseVLM \cite{zhang2025sparsevlm}, LLaVA-PruMerge \cite{shang2025llavaprumerge}, VisPruner \cite{zhang2025vispruner}, DivPrune \cite{alvar2025divprune}, TopV \cite{yang2025topv}, VScan \cite{zhang2025vscan}, and VisionZip \cite{yang2025visionzip}. Recent extensions to video VLMs further model temporal or spatiotemporal redundancy, including PruneVid \cite{huang2025prunevid}, FrameFusion \cite{fu2024framefusion}, FastVID \cite{shen2025fastvid}, DyCoke \cite{tao2025dycoke}, STTM \cite{hyun2025sttm}, HoliTom \cite{shao2025holitom}, DyToK \cite{li2025dytok}, and FlashVID \cite{fan2026flashvid}. These methods typically exploit spatial homogeneity or temporal continuity to filter out uninformative background patches or fuse visually correlated regions. However, they mainly treat compression as a plug-and-play acceleration module and focus on efficient inference. By contrast, our work does not propose a new compression operator; instead, it studies how a video VLM should be \emph{adapted} once compression is introduced into the training pipeline, with particular emphasis on mitigating the severe semantic drift that occurs during fine-tuning.

\paragraph{Parameter-efficient adaptation and distillation}
Parameter-efficient fine-tuning reduces adaptation cost by updating only a small subset of parameters rather than the full backbone. Representative approaches include adapters \cite{houlsby2019adapters,Chen_2026_CVPR,zhang20261st}, prefix tuning \cite{li2021prefix}, prompt tuning \cite{lester2021prompt}, LoRA \cite{hu2022lora}, AdaLoRA \cite{zhang2023adalora}, and QLoRA \cite{dettmers2023qlora}. Knowledge distillation provides a complementary route to efficient transfer, with representative methods including the original teacher-student framework \cite{hinton2015distilling}, DistilBERT \cite{sanh2019distilbert}, TinyBERT \cite{jiao2020tinybert}, and MiniLM \cite{wang2020minilm}. Existing PEFT and distillation methods, however, do not explicitly address the mismatch introduced by token-compressed video inputs. Crucially, standard distillation paradigms assume a one-to-one token correspondence or consistent feature manifolds between teacher and student, an assumption that collapses when inputs are aggressively compressed. 


\section{Method}
\subsection{Problem Setup}

We study the efficient adaptation of a video vision-language model (VLM) under a strict visual token budget. Let $\mathcal{D}=\{(v, q, y)\}$ denote the training set, where $v$ is a video, $q$ is the textual instruction or question, and $y=(y_1,\ldots,y_T)$ is the target answer sequence. In long-video scenarios, the number of visual tokens scales linearly with the number of frames, leading to a quadratic explosion in self-attention complexity that makes both fine-tuning and inference computationally prohibitive.

To mitigate this bottleneck, we consider a token-budgeted setting where the model is constrained to consume at most $B$ visual tokens per video sample. Given the full visual token sequence $\mathbf{z}^{f}=\{z_1,\ldots,z_{N_f}\}$ extracted from $v$, we apply a compression operator $\mathcal{C}(\cdot; B)$ to yield a compact sequence:
\begin{equation}
\mathbf{z}^{c} = \mathcal{C}(\mathbf{z}^{f}; B), \quad |\mathbf{z}^{c}| = N_c \leq B \ll N_f,
\end{equation}
where $\mathbf{z}^{f}$ and $\mathbf{z}^{c}$ denote the full and compressed token representations, respectively. In our implementation, $\mathcal{C}$ is instantiated via FlashVID-based spatiotemporal token compression along the video pathway. The fundamental challenge lies in preserving the fine-grained visual semantics critical for accurate reasoning when operating on this heavily compressed representation.

Our objective is to adapt the model to operate efficiently on $\mathbf{z}^{c}$ while strictly preserving the semantic reasoning capabilities of the full-token model. To achieve this, we freeze the pretrained backbone parameters $\Theta$ and optimize only a set of low-rank adaptation (LoRA) parameters $\Phi$. Formally, the optimization target is formulated as:
\begin{equation}
\Phi^{*} = \arg\min_{\Phi} \sum_{(v,q,y)\in\mathcal{D}} \mathcal{L}\big(y; v, q, \mathbf{z}^{f}, \mathbf{z}^{c}; \Theta, \Phi\big),
\end{equation}
where $\mathcal{L}$ elegantly combines the standard supervised objective with our proposed reliability-aware distillation constraints, as detailed in the subsequent subsections.

\subsection{Token-Budget Distillation Framework}

Building upon this formulation, we propose \emph{Token-Budget Distillation} (TBD), a dual-path training framework designed to seamlessly map full-token semantics into a compressed-token student (Figure~\ref{fig:pipeline}). For each training instance $(v,q,y)$, TBD constructs two distinct forward pathways from the identical pretrained backbone: a \emph{teacher} branch processing $\mathbf{z}^{f}$ and a \emph{student} branch processing $\mathbf{z}^{c}$.

The teacher branch operates with compression disabled and LoRA adapters deactivated, effectively acting as an exact proxy for the original full-token model. Let
\begin{equation}
\mathbf{L}^{t} = f(v,q;\mathbf{z}^{f}, \Theta), \qquad
\mathbf{L}^{s} = f(v,q;\mathbf{z}^{c}, \Theta, \Phi),
\end{equation}
denote the next-token logit sequences for the teacher and student, respectively. The corresponding probability distributions are obtained via the softmax operator $\sigma(\cdot)$. Crucially, $\mathbf{L}^{t}$ is generated in a gradient-free evaluation mode, whereas $\mathbf{L}^{s}$ is produced by the trainable compressed branch. 

This asymmetric architecture ensures that the teacher branch provides a robust, anchor-like full-token target, compelling the student to reconstruct this high-fidelity behavior from scarce visual evidence. Post-adaptation, the teacher branch is discarded, yielding a highly efficient student for inference.

\begin{algorithm}[htbp]
\caption{Token-Budget Distillation (TBD) Optimization}
\label{alg:tbd}
\textbf{Input:} Training set $\mathcal{D}$, frozen VLM backbone $\Theta$, trainable LoRA adapters $\Phi$, compression budget $B$, KD hyperparameters ($\lambda_{\mathrm{ce}}, \lambda_{\mathrm{m}}, \lambda_{\mathrm{kd}}$).\\
\textbf{Output:} Optimized student LoRA parameters $\Phi^{*}$
\begin{algorithmic}[1]
\FOR{each batch $(v, q, y) \in \mathcal{D}$}
    \STATE \textcolor{gray}{\# 1. Full-Token Teacher Forward (No Gradient)}
    \STATE Extract full visual token sequence $\mathbf{z}^{f}$ from $v$
    \STATE $\mathbf{L}^{t} \leftarrow f(v, q; \mathbf{z}^{f}, \Theta)$ \quad \textcolor{gray}{\# Compression disabled}
    
    \STATE \textcolor{gray}{\# 2. Compressed Student Forward (With Gradient)}
    \STATE $\mathbf{z}^{c} \leftarrow \mathcal{C}(\mathbf{z}^{f}; B)$ \quad \textcolor{gray}{\# FlashVID compression}
    \STATE $\mathbf{L}^{s} \leftarrow f(v, q; \mathbf{z}^{c}, \Theta, \Phi)$ \quad \textcolor{gray}{\# LoRA enabled}
    
    \STATE \textcolor{gray}{\# 3. Reliability Masking \& Alignment}
    \STATE Identify aligned answer positions $M = \min(M_t, M_s)$
    \STATE Initialize $\Omega \leftarrow \emptyset$
    \FOR{$i = 1$ \TO $M$}
        \IF{Teacher is correct and confident at step $i$}
            \STATE $\Omega \leftarrow \Omega \cup \{i\}$
        \ENDIF
    \ENDFOR
    
    \STATE \textcolor{gray}{\# 4. Objective Computation}
    \STATE Compute Task Loss: $\mathcal{L}_{\mathrm{CE}} \leftarrow \text{CrossEntropy}(\mathbf{L}^{s}, y)$
    \STATE Compute KL Alignment: $\mathcal{L}_{\mathrm{KL}} \leftarrow \text{KL-Div}(\mathbf{L}^{t}_{\Omega}, \mathbf{L}^{s}_{\Omega})$
    \STATE Compute Margin Gap: $\mathcal{L}_{\mathrm{margin}} \leftarrow \text{Huber}(\mathbf{m}^{s}_{\Omega}, \mathbf{m}^{t}_{\Omega})$
    \STATE Calculate Distillation Loss: $\mathcal{L}_{\mathrm{KD}} \leftarrow \mathcal{L}_{\mathrm{KL}} + \lambda_{\mathrm{m}} \mathcal{L}_{\mathrm{margin}}$
    
    \STATE \textcolor{gray}{\# 5. Dynamic Weighting \& Update}
    \STATE Update $\hat{\lambda}_{\mathrm{kd}}(t)$ via warmup and CE-relative capping
    \STATE $\mathcal{L} \leftarrow \lambda_{\mathrm{ce}} \mathcal{L}_{\mathrm{CE}} + \hat{\lambda}_{\mathrm{kd}}(t) \mathcal{L}_{\mathrm{KD}}$
    \STATE $\Phi \leftarrow \Phi - \eta \nabla_{\Phi} \mathcal{L}$
\ENDFOR
\end{algorithmic}
\end{algorithm}

\subsection{Compression-Aware LoRA Adaptation}

The student branch harmonizes visual token compression with parameter-efficient fine-tuning (PEFT). Specifically, FlashVID compresses the spatiotemporal token sequence $\mathbf{z}^{f}$ into $\mathbf{z}^{c}$ before it enters the core language model. We integrate FlashVID as a dynamic runtime module whose state (on/off) dictates whether the backbone acts as the student or teacher.

To enable adaptation under this compression, we freeze the backbone $\Theta$ and inject LoRA parameters $\Phi$. For any adaptable linear transformation with pretrained weights $W_0$, the updated weight is parameterized as:
\begin{equation}
W = W_0 + \Delta W, \qquad \Delta W = B_W A_W,
\end{equation}
where $A_W \in \mathbb{R}^{r \times d}$ and $B_W \in \mathbb{R}^{d' \times r}$ are trainable low-rank matrices ($r \ll \min(d,d')$). Compression and adaptation thus serve complementary purposes: FlashVID tackles the efficiency bottleneck by slashing the sequence length, while LoRA neutralizes the ensuing semantic drift via targeted residual updates. 

Notably, the token selection and merging mechanisms within FlashVID remain entirely training-free. Gradients only flow through the compressed representations to update the LoRA adapters, allowing TBD to dedicate its entire learning capacity to representation rehabilitation rather than optimizing the compression logic itself.

\subsection{Reliable Distillation Under Token Compression}

Compression-aware task tuning alone is insufficient to prevent semantic drift. We therefore distill the student's output toward the teacher's full-token predictions. However, token compression fundamentally alters the model's predictive certainty; thus, uniform distillation across all tokens introduces harmful noise. To combat this, TBD introduces \emph{reliable answer-region distillation}, explicitly isolating high-quality transfer signals.

Following standard causal language modeling, we shift the logits and identify valid answer positions. Let $\mathbf{l}^{t}_{i}$ and $\mathbf{l}^{s}_{i}$ represent the aligned teacher and student logits at position $i$. Given valid answer lengths $M_t$ and $M_s$, we define the alignment horizon as $M = \min(M_t, M_s)$. 

Within this aligned region, we compute a token-level KL divergence:
\begin{equation}
\mathcal{L}_{\mathrm{KL}}
=
\frac{1}{|\Omega|}
\sum_{i \in \Omega}
\mathrm{KL}\Big(
\sigma(\mathbf{l}^{t}_{i}/\tau)
\;\|\;
\sigma(\mathbf{l}^{s}_{i}/\tau)
\Big)\tau^2,
\end{equation}
where $\tau$ is the temperature. The critical innovation here is the construction of the reliability mask $\Omega$. We populate $\Omega$ exclusively with positions where the teacher is both \emph{correct} (highest probability assigned to the ground-truth) and \emph{confident} (probability exceeds a predefined threshold).

To strictly prevent margin collapse—a common symptom of semantic drift where the model struggles to differentiate between the correct concept and related background noise—we introduce \emph{GT-anchored margin distillation}. For each $i \in \Omega$, let $y_i$ be the ground-truth token and $n_i$ be the hardest negative token (the highest-scoring incorrect token per the teacher). We explicitly formalize their margins:
\begin{equation}
m^{t}_{i} = l^{t}_{i,y_i} - l^{t}_{i,n_i},
\qquad
m^{s}_{i} = l^{s}_{i,y_i} - l^{s}_{i,n_i},
\end{equation}
and penalize deviations using a robust Huber loss:
\begin{equation}
\mathcal{L}_{\mathrm{margin}}
=
\frac{1}{|\Omega|}
\sum_{i \in \Omega}
\mathrm{Huber}(m^{s}_{i}, m^{t}_{i}).
\end{equation}
This ensures the compressed student mimics not just the softened distribution, but the teacher's decisive boundary gaps.

\subsection{Overall Objective and Optimization}

The complete optimization objective balances standard autoregressive training with our reliability-aware distillation. The primary task loss ensures fidelity to the downstream instruction:
\begin{equation}
\mathcal{L}_{\mathrm{CE}}
=
- \frac{1}{|\mathcal{A}|}\sum_{i \in \mathcal{A}} \log \sigma(\mathbf{l}^{s}_{i})_{y_i},
\end{equation}
where $\mathcal{A}$ spans all valid student answer positions.

The distillation objective aggregates the KL and margin terms:
\begin{equation}
\mathcal{L}_{\mathrm{KD}}
=
\mathcal{L}_{\mathrm{KL}}
+
\lambda_{\mathrm{m}} \mathcal{L}_{\mathrm{margin}}.
\end{equation}

The final training loss is dynamically scaled:
\begin{equation}
\mathcal{L}
=
\lambda_{\mathrm{ce}} \mathcal{L}_{\mathrm{CE}}
+
\hat{\lambda}_{\mathrm{kd}}(t)\,\mathcal{L}_{\mathrm{KD}}.
\end{equation}
To prevent early-stage over-regularization, $\hat{\lambda}_{\mathrm{kd}}(t)$ incorporates a linear warmup schedule and is strictly capped relative to the magnitude of $\mathcal{L}_{\mathrm{CE}}$. The complete training pipeline is summarized in Algorithm~\ref{alg:tbd}.

\begin{table*}[t]
\centering
\scriptsize
\setlength{\tabcolsep}{2pt}
\renewcommand{\arraystretch}{1.0}
\caption{Main results on four video understanding benchmarks under different token retention ratios. We compare TBD with the uncompressed Vanilla model and representative token reduction baselines on three backbones: \textbf{LLaVA-Video}, \textbf{LLaVA-OneVision}, and \textbf{Qwen3-VL-8B-Instruct}. ``Avg. Score'' denotes the average performance across benchmarks, and ``Rel. Acc (\%)'' reports the average score normalized by the corresponding Vanilla model.}
\label{tab:main_results}
\resizebox{1.0 \textwidth}{!}{
\begin{tabular}{lccccccccccc}
\toprule
\multirow{2}{*}{\textbf{Method}} 
& \multirow{2}{*}{\makecell{\textbf{Retention} \\ \textbf{Ratio} $R$}}
& \multicolumn{4}{c}{\textbf{VideoMME}} 
& \multicolumn{2}{c}{\textbf{EgoSchema}} 
& \multirow{2}{*}{\makecell{\textbf{LongVideo} \\ \textbf{Bench}}} 
& \multirow{2}{*}{\textbf{MVBench}} 
& \multicolumn{2}{c}{\textbf{Avg.}} \\

\cmidrule(lr){3-6}
\cmidrule(lr){7-8}
\cmidrule(lr){11-12}

& & \textbf{Short} & \textbf{Medium} & \textbf{Long} & \textbf{Overall}
& \textbf{Subset} & \textbf{Total} & & & \textbf{Score} & \textbf{Rel. Acc (\%)} \\
\midrule

\rowcolor{green!12}
\multicolumn{12}{c}{\textbf{LLaVA-Video}} \\
{Vanilla} 
    & 100\% & 77.0 & 62.1 & 53.3 & 64.2 & 59.4 & 57.3 & 59.5 & 61.9 & 60.7 & 100.0\\ 
             
\midrule
{FastV} 
    & \multirow{5}{*}{\makecell{20\%}} & 69.3 & 58.3 & 49.9 & 59.2 & 54.8 & 54.1 & 56.0 & 58.4 & 56.9 & 93.7\\ 

{VisionZip} 
    &  & 72.3 & 59.6 & \textbf{53.3} & 61.7 & \textbf{59.0} & 56.4 & 58.0 & 59.8 & 59.0 & 97.2\\ 

{FastVID} 
    &  & \textbf{74.6} & 60.8 & 52.3 & 62.6 & 57.0 & 55.0 & 57.1 & 60.2 & 58.7 & 96.7\\ 

{FlashVID} 
    &  & 74.1 & 60.0 & 52.3 & 62.2 & 58.4 & 56.4 & 58.7 & 59.8 & 59.3 & 97.7\\ 

{TBD (ours)} 
    &  & 74.0 & \textbf{61.0} & 53.1 & \textbf{62.7} & 58.5 & \textbf{56.6} & \textbf{59.2} & \textbf{60.3} & \textbf{59.7} & \textbf{98.4}\\ 
\midrule
{FastV} 
    & \multirow{5}{*}{\makecell{10\%}} & 64.3 & 53.8 & 49.2 & 55.8 & 50.6 & 51.1 & 53.6 & 56.2 & 54.2 & 89.3\\ 

{VisionZip} 
    &  & 69.4 & 57.9 & 51.2 & 59.5 & 54.4 & 53.9 & 54.5 & 58.5 & 56.6 & 93.2\\ 

{FastVID} 
    &  & 71.8 & 57.3 & 50.2 & 59.8 & 54.8 & 52.4 & 56.9 & 59.3 & 57.1 & 94.1\\ 

{FlashVID} 
    &  & 72.2 & 59.1 & 51.2 & 60.9 & 57.2 & 54.9 & 57.7 & 59.3 & 58.2 & 95.9\\ 

{TBD (ours)} 
    &  & \textbf{73.6} & \textbf{60.4} & \textbf{52.6} & \textbf{62.2} & \textbf{57.4} & \textbf{55.2} & \textbf{58.1} & \textbf{60.0} & \textbf{58.9} & \textbf{97.0}\\ 
                                    
\midrule

\rowcolor{blue!12}
\multicolumn{12}{c}{\textbf{LLaVA-OneVision}} \\
{Vanilla} 
    & 100\% & 69.9 & 56.7 & 48.9 & 58.5 & 62.2 & 60.3 & 56.6 & 58.3 & 58.4 & 100.0\\            
\midrule
{FastV} 
    & \multirow{5}{*}{\makecell{20\%}} & 66.3 & 53.9 & 46.9 & 55.7 & 60.6 & 57.6 & 56.0 & 56.0 & 56.3 & 96.4\\ 
         
{VisionZip} 
    &  & 68.6 & \textbf{57.0} & 48.3 & 58.0 & 62.0 & 60.0 & 55.4 & 57.6 & 57.7 & 98.8\\ 

{FastVID} 
    &  & 69.9 & 56.3 & 47.4 & 57.9 & 61.2 & 59.5 & 55.9 & 58.1 & 57.9 & 99.1\\ 

{FlashVID} 
    &  & 70.1 & 55.4 & 48.9 & 58.2 & \textbf{63.0} & 60.1 & 58.5 & 58.2 & 58.7 & 100.5\\ 

{TBD (ours)} 
    &  & \textbf{71.3} & 56.8 & \textbf{49.1} & \textbf{59.1} & 62.9 & \textbf{60.1} & \textbf{58.6} & \textbf{58.3} & \textbf{59.0} & \textbf{101.0}\\ 
\midrule
{FastV} 
    & \multirow{5}{*}{\makecell{10\%}} & 60.9 & 52.2 & 44.9 & 52.7 & 59.0 & 56.0 & 52.4 & 53.4 & 53.6 & 91.8\\ 

{VisionZip} 
    &  & 60.3 & 52.9 & 46.7 & 53.3 & 61.6 & 58.5 & 49.4 & 54.8 & 54.0 & 92.5\\ 

{FastVID} 
    &  & 68.1 & 55.7 & 47.8 & 57.2 & 58.8 & 58.7 & 55.7 & 57.0 & 57.1 & 97.8\\ 

{FlashVID} 
    &  & 67.3 & 57.1 & 49.0 & 57.8 & 62.4 & 60.0 & 56.5 & 57.4 & 57.9 & 99.1\\

{TBD (ours)} 
    &  & \textbf{68.5} & \textbf{58.4} & \textbf{50.2} & \textbf{59.0} & \textbf{62.4} & \textbf{60.1} & \textbf{56.9} & \textbf{57.7} & \textbf{58.4} & \textbf{100.0}\\  

\midrule

\rowcolor{yellow!12}
\multicolumn{12}{c}{\textbf{Qwen3-VL-8B-Instruct}} \\
{Vanilla} 
    & 100\% & 74.6 & 60.8 & 56.3 & 63.9 & 66.2 & 68.3 & 59.2 & 66.3 & 64.4 & 100.0\\            
\midrule
{FastVID} 
    & \multirow{3}{*}{\makecell{20\%}} & 66.9 & 54.1 & 51.8 & 57.6 & 62.2 & 62.0 & 56.7 & 63.3 & 59.9 & 93.0\\ 

{FlashVID} 
    &  & 68.3 & 55.3 & 52.1 & 58.6 & 64.4 & 63.2 & 57.2 & 64.6 & 60.9 & 94.6\\ 

{TBD (ours)} 
    & & \textbf{69.2} & \textbf{56.4} & \textbf{53.0} & \textbf{59.5} & \textbf{65.0} & \textbf{63.8} & \textbf{57.9} & \textbf{65.1} & \textbf{61.6} & \textbf{95.7} \\
\midrule
{FastVID} 
    & \multirow{3}{*}{\makecell{10\%}} & 64.6 & 54.5 & 49.3 & 56.1 & 61.8 & 61.4 & 54.6 & 58.9 & 57.8 & 89.8\\ 
    
{FlashVID} 
    &  & 66.1 & 55.3 & 49.8 & 57.1 & 63.0 & 62.2 & 55.2 & 60.5 & 58.8 & 91.3\\

{TBD (ours)} 
    &  & \textbf{67.3} & \textbf{56.3} & \textbf{50.6} & \textbf{58.1} & \textbf{63.5} & \textbf{62.6} & \textbf{55.8} & \textbf{61.0} & \textbf{59.4} & \textbf{92.2}\\

\bottomrule
\end{tabular}
}
\end{table*}

\section{Experiments}
\subsection{Experimental Setup}
\paragraph{Training Data and Details}
We train our models strictly on the comprehensive video instruction-tuning dataset LLaVA-Video-178K \cite{zhang2024llava}. This dataset is specifically chosen for its high-quality, diverse spatiotemporal annotations, ensuring that the teacher model's semantic guidance during our distillation process is rooted in complex, multi-turn reasoning rather than superficial visual matching. By utilizing a rich instruction-following corpus, we guarantee that the adapted student model learns to align compressed visual tokens with complex linguistic queries.
\paragraph{Evaluation Benchmarks}
To rigorously assess the generalizability and robustness of TBD, we evaluate on four distinct and highly challenging video understanding benchmarks. Specifically, \textbf{VideoMME} \cite{fu2025video} provides a comprehensive evaluation across short, medium, and long videos, testing the model's versatility; \textbf{EgoSchema} \cite{mangalam2023egoschema} challenges the model with extremely long-term, goal-oriented reasoning in egocentric videos; \textbf{LongVideoBench} \cite{wu2024longvideobench} focuses on extended contextual memory and understanding; and \textbf{MVBench} \cite{li2024mvbench} probes fine-grained spatiotemporal dynamics and micro-action recognition. By covering both macro-level narrative comprehension and micro-level action details, this evaluation suite ensures that any semantic drift, temporal hallucination, or boundary collapse caused by token compression is thoroughly exposed and quantified. Following standard protocols, we report benchmark-specific accuracy metrics and additionally summarize the average performance across all tasks to reflect the overarching accuracy-efficiency trade-off under strictly constrained token budgets.
\paragraph{Model Backbones}
We instantiate and evaluate TBD across three highly representative, state-of-the-art video VLM backbones: LLaVA-Video \cite{zhang2024llava}, LLaVA-OneVision \cite{li2024llava}, and Qwen3-VL-8B-Instruct \cite{bai2025qwen3}. These backbones span different architectural paradigms, spatial-temporal pooling strategies, and parameter scales. Evaluating across this diverse set allows us to demonstrate the universal applicability and plug-and-play nature of our distillation framework, proving that TBD's effectiveness is not strictly tied to a specific visual encoder or LLM family.
\paragraph{Baselines and Compression Regimes}
In the main results, we compare TBD against the original uncompressed model (Vanilla, serving as the theoretical upper bound) and a suite of representative token reduction baselines. These baselines encompass both spatial-only token dropping/merging (FastV \cite{chen2024image}, VisionZip \cite{yang2025visionzip}) and advanced spatiotemporal reduction techniques (FastVID \cite{shen2025fastvid}, FlashVID \cite{fan2026flashvid}), representing the current landscape of training-free visual compression. For a strictly fair comparison, all compressed variants—whether training-free or adapted—are evaluated under identical visual token retention ratios. We report results at $R=20\%$ to examine performance under a moderate efficiency bottleneck, and at a highly aggressive $R=10\%$ regime to explicitly stress-test the model's ability to preserve cognitive reasoning when $90\%$ of the visual evidence is permanently discarded.

\subsection{Main Results}

Table~\ref{tab:main_results} summarizes the main results. TBD consistently improves the accuracy-efficiency trade-off of compressed video VLMs across three backbones and different compression regimes. Compared with compression-only baselines, TBD achieves the strongest or near-strongest performance on most benchmarks while operating under the same token budget. This trend indicates that the proposed distillation-based adaptation is effective at recovering the semantic information lost during token compression, especially when the retention ratio becomes more aggressive.

On LLaVA-Video, TBD achieves the best average score among compressed variants at both $R=20\%$ and $R=10\%$. At $R=20\%$, it reaches an average score of 59.7, outperforming FlashVID (59.3), FastVID (58.7), VisionZip (59.0), and FastV (56.9), while retaining 98.4\% of the Vanilla model's average accuracy. At the more aggressive $R=10\%$ setting, TBD further shows clear robustness, obtaining an average score of 58.9 and preserving 97.0\% of the Vanilla accuracy, again surpassing all compared compressed baselines. We also observe consistent gains on the overall VideoMME score, indicating that TBD remains effective across short, medium, and long video subsets.
Similar trends hold for LLaVA-OneVision. At $R=20\%$, TBD achieves the highest average score among compressed methods (59.0), slightly surpassing FlashVID (58.7) and even marginally exceeding the Vanilla model in relative accuracy. At $R=10\%$, TBD remains the strongest compressed variant with an average score of 58.4, improving over FlashVID (57.9), FastVID (57.1), VisionZip (54.0), and FastV (53.6), while matching the Vanilla model in relative average accuracy.
We further evaluate TBD on Qwen3-VL-8B-Instruct. Compared with FlashVID, TBD consistently improves performance at both retention ratios. At $R=20\%$, the average score increases from 60.9 to 61.6, with gains on all four benchmarks. At the more aggressive $R=10\%$ setting, TBD improves the average score from 58.8 to 59.4 and raises the relative accuracy from 91.3\% to 92.2\%. These results show that the benefit of TBD is not limited to the LLaVA family, but also generalizes to a stronger and architecturally different video VLM backbone.

Overall, these results confirm that TBD is effective across different video VLM backbones and retains its advantage under both moderate and aggressive token compression.

\subsection{Ablation Study}

We conduct ablation experiments on LLaVA-Video under the more challenging retention ratio $R=10\%$ to quantify the contribution of each component in TBD. As shown in Table~\ref{tab:ablation}, the LoRA-only Student baseline achieves an average score of 58.3, indicating that parameter-efficient adaptation alone can partially compensate for compression, but still leaves a clear gap to the full method.

Adding answer-region KL distillation improves the average score from 58.3 to 58.6, showing that aligning the compressed student with the full-token teacher at the output-distribution level is already beneficial. When we further incorporate GT-anchored margin distillation, the score increases to 58.8, suggesting that explicitly preserving the teacher's local decision boundary provides additional gains beyond distribution matching alone.

Finally, enabling the full set of reliability filtering and CE-relative KD capping yields the best result, reaching 58.9 average score and 97.0\% relative accuracy. Although the improvement over the previous variant is smaller, it is consistent with the role of these mechanisms: they primarily stabilize training and suppress noisy distillation signals, thereby making the overall optimization more reliable. Overall, the ablation study confirms that the gains of TBD come from the complementary effects of answer-region distribution alignment, margin-based boundary preservation, and reliability-aware KD control.
\begin{table}[t]
\centering
\caption{Ablation study on \textbf{LLaVA-Video} at retention ratio $R=10\%$. ``Rel. Acc (\%)'' is normalized by the corresponding Vanilla model in Table~\ref{tab:main_results}.}
\label{tab:ablation}
\resizebox{\columnwidth}{!}{%
\renewcommand{\arraystretch}{1.05}
\begin{tabular}{lcccccc}
\toprule
\textbf{Variant} & \textbf{LoRA} & \makecell{\textbf{Answer-region} \\ \textbf{KL}} & \makecell{\textbf{GT-anchored} \\ \textbf{Margin}} & \makecell{\textbf{Filtering} \\ \textbf{\& KD Cap}} & \textbf{Score} & \textbf{Rel. Acc (\%)} \\
\midrule
LoRA-only Student & \cmark & \xmark & \xmark & \xmark & 58.3 & 96.0 \\
 + Answer-region KL & \cmark & \cmark & \xmark & \xmark & 58.6 & 96.5 \\
 + GT-anchored Margin & \cmark & \cmark & \cmark & \xmark & 58.8 & 96.9 \\
Full TBD & \cmark & \cmark & \cmark & \cmark & 58.9 & 97.0 \\
\bottomrule
\end{tabular}%
}
\end{table}

\subsection{Effect of Compression Modules}

We further examine the sensitivity of TBD to the choice of compression module by replacing FlashVID with three alternative token compression methods, namely FastVID, VisionZip, and FastV, while keeping the same TBD training objective. All comparisons are conducted on LLaVA-Video at retention ratio $R=10\%$.

As shown in Table~\ref{tab:compression_modules}, TBD remains effective across different compression front-ends, but the choice of compression module still has a clear impact on the final performance. Among the four variants, TBD (FlashVID) achieves the best result, reaching an average score of 58.9 and 97.0\% relative accuracy, followed by TBD+FastVID (57.7, 95.1\%), TBD+VisionZip (57.1, 94.1\%), and TBD+FastV (55.0, 90.6\%). This trend is consistent with the relative strength of these compression methods in the main results.

These results suggest that TBD is not tied to a specific compression mechanism and can be combined with different token reduction strategies. At the same time, the performance gap across compression modules indicates that reliable distillation does not fully remove the influence of compression quality: a stronger compression front-end still provides a better basis for preserving full-token semantics under the same token budget.

\begin{table}[t]
\centering
\small
\setlength{\tabcolsep}{8pt}
\renewcommand{\arraystretch}{1.05}
\caption{Effect of different compression modules within TBD on \textbf{LLaVA-Video} at retention ratio $R=10\%$. FlashVID is the default compression module used in TBD unless otherwise specified. ``Rel. Acc (\%)'' is normalized by the corresponding Vanilla model in Table~\ref{tab:main_results}.}
\label{tab:compression_modules}
\begin{tabular*}{\columnwidth}{@{\extracolsep{\fill}}lcc@{}}
\toprule
\textbf{Variant} & \textbf{Score} & \textbf{Rel. Acc (\%)} \\
\midrule
TBD+FastV & 55.0 & 90.6 \\
TBD+VisionZip & 57.1 & 94.1 \\
TBD+FastVID & 57.7 & 95.1 \\
\textbf{TBD (FlashVID)} & \textbf{58.9} & \textbf{97.0} \\
\bottomrule
\end{tabular*}
\end{table}

\subsection{Logit Distribution Alignment Analysis}
To further examine whether TBD improves output-level alignment with the full-token teacher, we analyze next-token logits on VideoMME. Specifically, we randomly sample 100 evaluation instances and compare three models under the same setting: the full-token Teacher, a compression-only baseline, and the distilled student trained with TBD. All compressed-model analyses are conducted on LLaVA-Video with retention ratio $R=10\%$.

\begin{figure}[h]
  \centering
  \includegraphics[width=\linewidth]{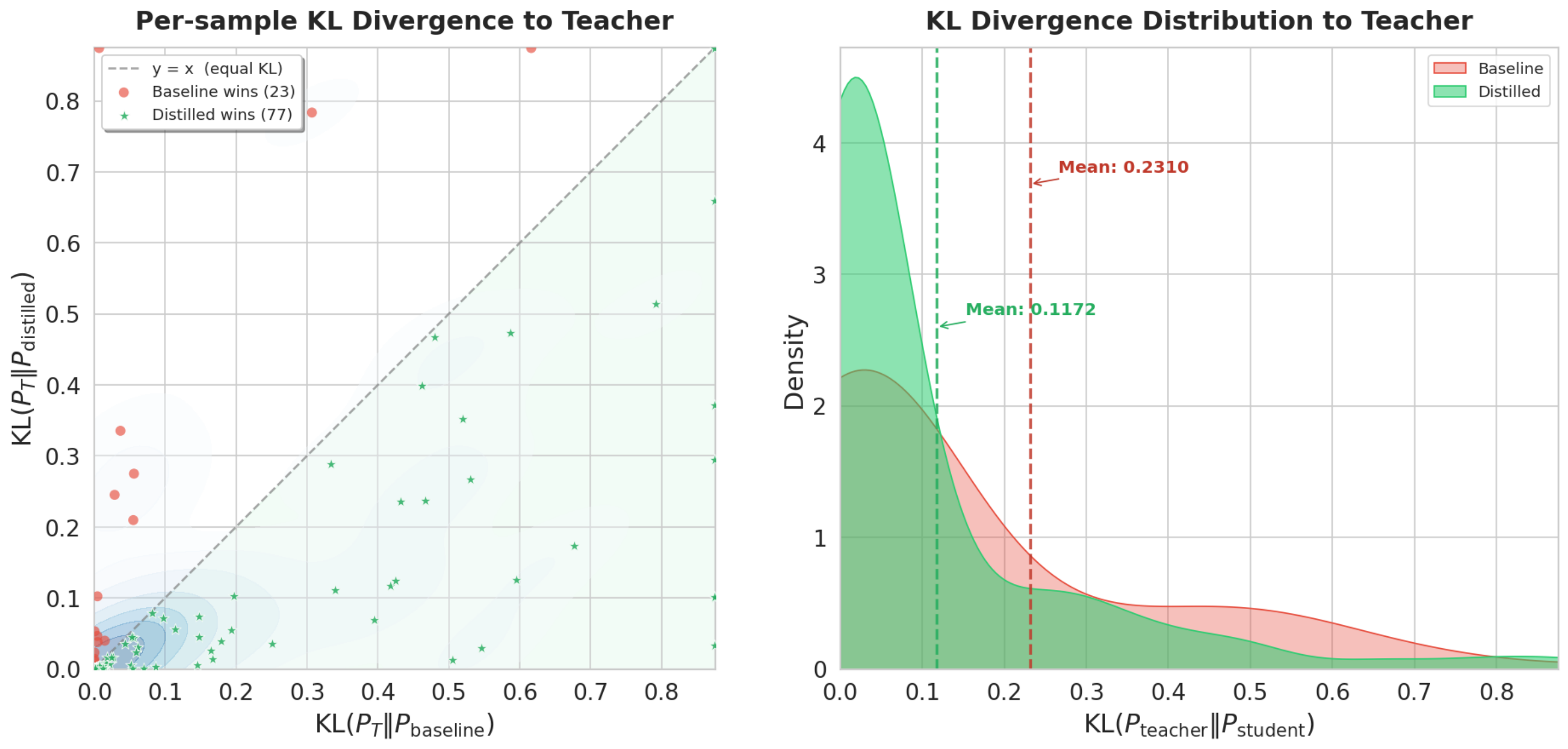}
  \caption{Logit distribution alignment on 100 randomly sampled VideoMME instances for \textbf{LLaVA-Video} at $R=10\%$. Left: per-sample comparison of $\mathrm{KL}(P_T \| P_{\mathrm{baseline}})$ and $\mathrm{KL}(P_T \| P_{\mathrm{distilled}})$ on normalized A/B/C/D choice probabilities. Right: kernel density estimates of the KL distributions, where the distilled student shows a substantially smaller mean KL divergence to the teacher.}
  \Description{Two-panel comparison of output alignment with the teacher. The left panel is a scatter plot in which each point is one sample, plotting the baseline's KL divergence against the distilled student's KL divergence; most points lie below the diagonal, meaning the student is closer to the teacher. The right panel shows density curves of the two KL distributions, with the student's distribution concentrated at clearly smaller values.}
  \label{fig:kl_divergence}
\end{figure}

For each sample, we extract the next-token logits, keep only the four answer-choice tokens $\{A,B,C,D\}$, normalize them with a softmax over the four candidates, and compute the per-sample KL divergence from the teacher distribution to each compressed model. This yields $\mathrm{KL}(P_T \| P_{\mathrm{baseline}})$ and $\mathrm{KL}(P_T \| P_{\mathrm{distilled}})$ for every sample.
Figure~\ref{fig:kl_divergence} visualizes the resulting alignment behavior. In the left panel, each point corresponds to one VideoMME sample, plotted by its KL divergence to the teacher for the compression-only baseline and the distilled student, respectively. Points below the diagonal indicate cases where the distilled student is closer to the teacher. We observe that this occurs for 77 out of 100 samples, whereas the compression-only baseline is closer on only 23 samples. The right panel shows the corresponding KL distributions, where the mean KL divergence is reduced from 0.2310 for the compression-only baseline to 0.1172 for the distilled student.
These results provide direct evidence for the effect of answer-region distillation. Beyond improving benchmark accuracy, TBD makes the distilled student substantially closer to the full-token teacher in the answer-choice probability space. This observation is consistent with the gains in Table~\ref{tab:main_results} and supports our claim that the proposed distillation framework recovers teacher semantics under a strict token budget.

\subsection{Decision Boundary Analysis}

\begin{figure}[h]
    \centering
    \includegraphics[width=\linewidth]{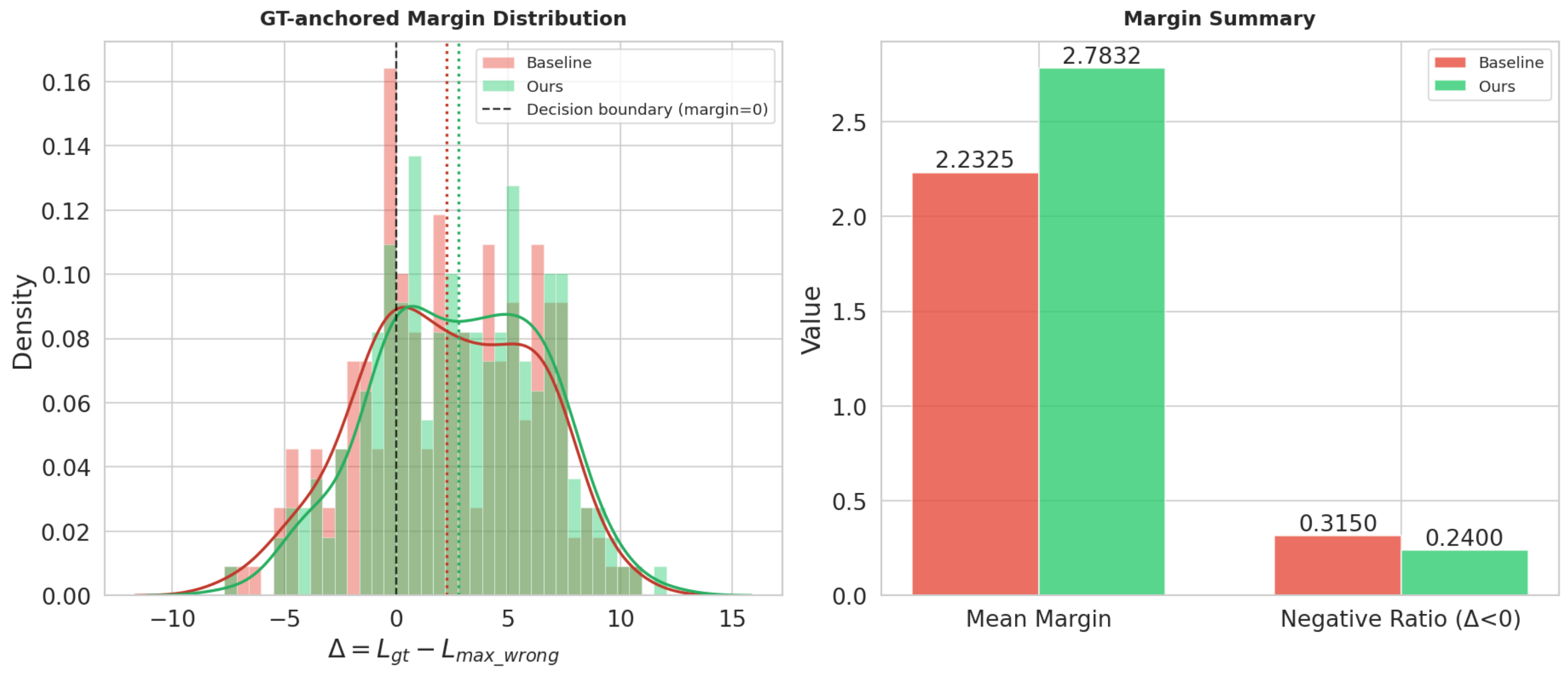}
    \caption{Decision boundary analysis on 200 randomly sampled VideoMME instances for \textbf{LLaVA-Video} at $R=10\%$. Left: histogram and density of the GT-anchored margin $\Delta = L_{gt} - L_{max\_wrong}$ for the compression-only baseline and the distilled student. Right: summary statistics showing that the distilled student achieves a larger mean margin and a lower negative-margin ratio.}
    \Description{Two-panel decision boundary comparison. The left panel overlays histograms and density curves of the ground-truth-anchored logit margin for the compression-only baseline and the distilled student, where the student's distribution is shifted toward larger positive margins. The right panel lists summary statistics showing the student attains a higher mean margin and a lower fraction of negative margins than the baseline.}
    \label{fig:margin_hist}
\end{figure}
\begin{figure}[h]
  \centering
  \includegraphics[width=\linewidth]{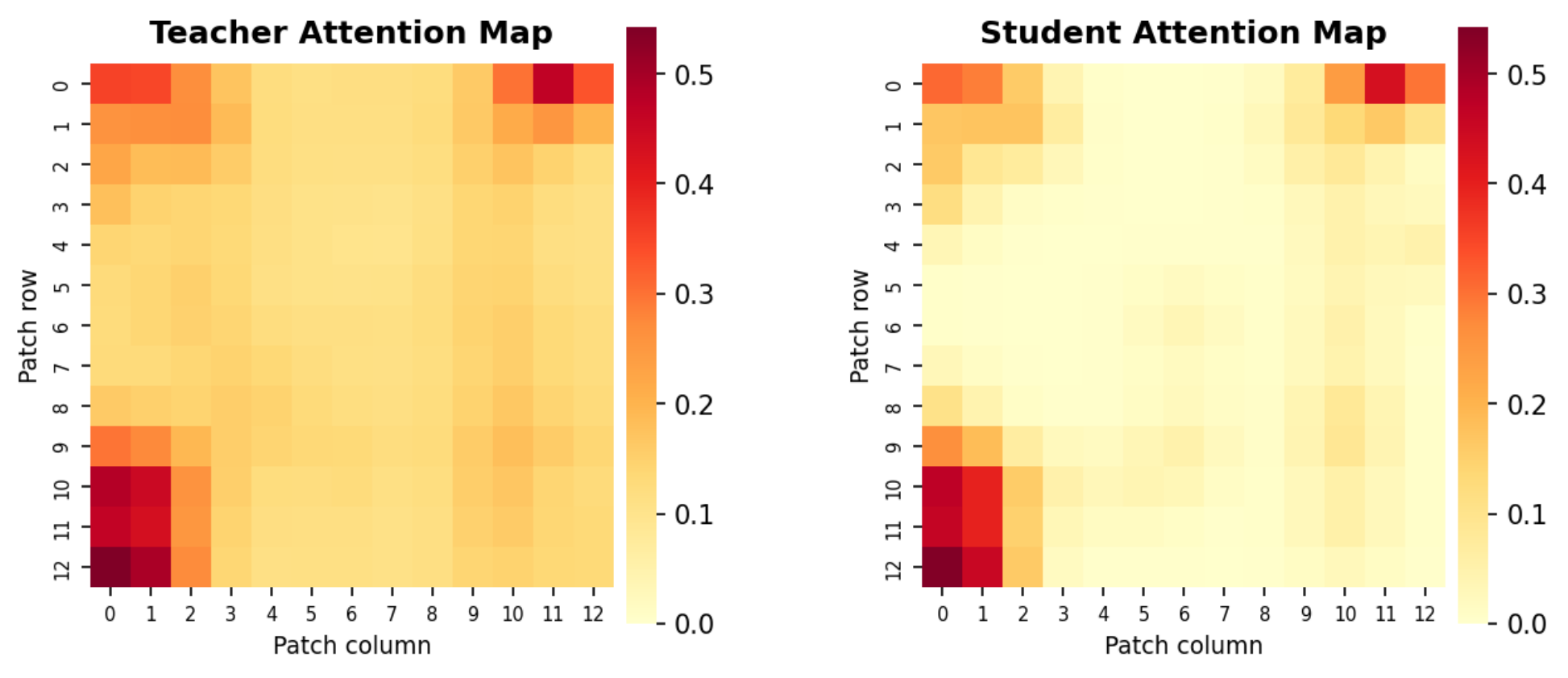}
  \caption{Attention drift visualization on a randomly selected \textbf{VideoMME} example for \textbf{LLaVA-Video} at retention ratio $R=10\%$. We compare the teacher and distilled student attention maps on the same frame.}
  \Description{Side-by-side spatial attention heatmaps over the same video frame for the full-token teacher and the TBD-adapted compressed student. Both heatmaps concentrate on the same task-relevant foreground region of the frame, showing that the student's visual attention stays consistent with the teacher despite aggressive token compression.}
  \label{fig:attn_drift}
\end{figure}

We next evaluate if GT-anchored margin distillation improves the decision boundary. We sample 200 VideoMME instances and compute the margin $\Delta = L_{gt} - L_{max\_wrong}$, where $L_{gt}$ is the ground-truth logit and $L_{max\_wrong}$ is the highest incorrect logit. A positive margin indicates confident correct predictions, while a negative one implies an incorrect preference.
Figure~\ref{fig:margin_hist} compares the compression-only baseline and our distilled student (LLaVA-Video, $R=10\%$). The distilled student's distribution explicitly shifts rightward (left panel), reflecting a stronger preference for the correct answer. Quantitatively (right panel), the mean margin increases from 2.2325 to 2.7832, and the negative-margin ratio drops from 0.3150 to 0.2400. This confirms the distilled student better separates the correct answer from the hardest distractor and makes fewer ranking errors.

Consistent with Section~3.4, while KL distillation aligns overall distributions, the margin term explicitly sharpens local decision boundaries. This mechanistically explains why the full TBD objective outperforms weaker variants in Table~\ref{tab:ablation}.

\subsection{Attention Drift Analysis}

To mechanistically understand how token compression impacts the model's internal reasoning, and to qualitatively verify that TBD successfully arrests semantic drift, we conduct an in-depth visual attention analysis. In standard video VLMs, the indiscriminate removal of spatiotemporal tokens often destroys fine-grained spatial granularity. Consequently, the self-attention mechanism becomes disoriented, leading to a phenomenon where the model's visual focus drifts away from critical foreground actions and disperses into irrelevant background noise or generalized textures. To examine whether our method mitigates this pathology, we visualize and compare the spatial attention maps of the full-token Teacher and the TBD-adapted Student. We extract these maps using the LLaVA-Video backbone evaluated on the VideoMME benchmark under a highly aggressive retention ratio of $R=10\%$. 
We randomly sample a representative video instance and select a keyframe to visualize the spatial attention distribution allocated to the visual tokens when generating the correct target answer. As vividly illustrated in Figure~\ref{fig:attn_drift}, the full-token Teacher acts as a reliable oracle, exhibiting a sharp, highly localized attention pattern that accurately fixates on the task-relevant foreground entities (e.g., the core action or specific object required to answer the query). Remarkably, despite operating on an aggressively truncated visual sequence where 90\% of the original tokens have been discarded, the TBD-adapted student's attention map remains highly consistent with that of the teacher. Rather than collapsing into a diffuse or shifted attention distribution—a typical symptom of naive compression—the TBD student successfully allocates its severely limited token budget to the exact same salient regions identified by the teacher.
This visualization yields a crucial insight: TBD does not merely act as a superficial output-level regularizer that forces logit matching. Instead, the combination of answer-region KL alignment and GT-anchored margin distillation fundamentally rehabilitates the internal multimodal routing of the network. By forcing the compressed student to reproduce the teacher's confident decision boundaries, TBD implicitly guides the student's attention heads to focus on the correct visual evidence, ensuring high efficiency without sacrificing interpretability and visual grounding.

\subsection{Discussion}
TBD reframes token-compressed adaptation as teacher-guided semantics
recovery rather than plain task fitting, and its advantage grows as
compression becomes more aggressive. The ablations confirm that the
proposed mechanisms are complementary: answer-region KL distillation
aligns output distributions, GT-anchored margin distillation preserves
discrimination, and reliability-aware filtering stabilizes noisy
signals, while the compression-module study shows TBD generalizes
across front-ends, with its upper bound set by front-end quality.
Mechanistically, these effects appear at multiple levels: the student
matches the teacher's answer-space distribution
(Figure~\ref{fig:kl_divergence}), sharpens decision boundaries around
the ground truth (Figure~\ref{fig:margin_hist}), and retains faithful
visual focus (Figure~\ref{fig:attn_drift}), which together explain the
robust gains.

\section{Conclusion and Limitations}
We presented \emph{Token-Budget Distillation} (TBD), a parameter-efficient framework for adapting video VLMs under strict token budgets. By synergizing token compression, LoRA, and reliable distillation, TBD enables compressed students to recover full-token semantic behavior. Extensive experiments across multiple benchmarks and three representative backbones demonstrate that TBD consistently improves the accuracy-efficiency trade-off over compression-only baselines, especially under aggressive compression. Ultimately, our findings suggest that effective adaptation under token compression is not solely an efficiency challenge, but a semantics preservation problem.
As for limitations, despite its efficacy at inference, TBD requires executing a full-token teacher during the adaptation phase, which maintains high training-time memory demands. Additionally, the student's performance upper bound is inherently constrained by the irreversible information loss of the chosen training-free compression module. We hope TBD provides a robust foundation for future research into fully memory-efficient and end-to-end compressed video VLMs.

\begin{acks}
This work was supported in part by the National Natural Science Foundation of China (NSFC) under Grant 62276283, in part by the China Meteorological Administration's Science and Technology Project under Grant CMAJBGS202517, in part by Guangdong-Hong Kong-Macao Greater Bay Area Meteorological Technology Collaborative Research Project under Grant GHMA2024Z04, in part by Fundamental Research Funds for the Central Universities, Sun Yat-sen University under Grant 23hytd006 and 23hytd006-2, and in part by Guangdong Provincial High-Level Young Talent Program under Grant RL2024-151-2-11.
\end{acks}

\bibliographystyle{ACM-Reference-Format}
\balance
\bibliography{references}

\end{document}